\documentclass{article}
\usepackage{spconf,amsmath,graphicx}
\usepackage{bm}
\usepackage{url}
\usepackage{fancyvrb}
\usepackage{bbding}
\usepackage{amssymb}
\usepackage{tipa}

\DeclareMathOperator*{\argmax}{arg\,max}
\DeclareMathOperator*{\mask}{\rm Mask}

\title{PHONEME-AWARE ENCODING FOR PREFIX-TREE-BASED CONTEXTUAL ASR}

\twoauthors
 {\begin{tabular}{c} Hayato Futami, Emiru Tsunoo, \\
  Yosuke Kashiwagi, Hiroaki Ogawa \end{tabular}}
	{Sony Group Corporation}
 {Siddhant Arora, Shinji Watanabe}
	{Carnegie Mellon University}

\begin{document}
\ninept
\maketitle

\begin{abstract}
In speech recognition applications, it is important to recognize context-specific rare words, such as proper nouns.
Tree-constrained Pointer Generator (TCPGen) has shown promise for this purpose, which efficiently biases such words with a prefix tree.
While the original TCPGen relies on grapheme-based encoding, we propose extending it with phoneme-aware encoding to better recognize words of unusual pronunciations.
As TCPGen handles biasing words as subword units, we propose obtaining subword-level phoneme-aware encoding by using alignment between phonemes and subwords.
Furthermore, we propose injecting phoneme-level predictions from CTC into queries of TCPGen so that the model better interprets the phoneme-aware encodings.
We conducted ASR experiments with TCPGen for RNN transducer.
We observed that proposed phoneme-aware encoding outperformed ordinary grapheme-based encoding on both the English LibriSpeech and Japanese CSJ datasets, demonstrating the robustness of our approach across linguistically diverse languages.
\end{abstract}
\section{Introduction}
\vspace{-5pt}
End-to-end automatic speech recognition (ASR) models have achieved excellent performances on general speech corpora \cite{Prabhavalkar23-E2E}.
However, they have difficulties in recognizing uncommon words such as names of persons and places, which can be crucial for downstream language understanding tasks.
Humans can estimate what those words are by considering which context they are in.
Contextual biasing is a method to incorporate such contextual knowledge into an end-to-end ASR model.
We pass the model a list of words that are likely to appear in the context and encourage the model to output these words, which is called biasing words \footnote{A list of phrases can be used for this. In this study, we assume words.}.

Many approaches have been considered for contextual biasing.
One approach is shallow fusion with a contextual language model (LM), where biasing words are compiled into Weight Finite State Transducer \cite{Zhao19-SF, Gourav21-PS}.
However, they require some heuristics and careful tuning of an LM weight to avoid under- or over-biasing, as they are not jointly optimized with ASR.
Meanwhile, attention-based deep context approaches \cite{Pundak18-DC, Jain20-CR, Sathyendra22-CA} have been considered to realize joint optimization with end-to-end ASR.
In these approaches, each biasing word is converted into an encoding vector, and an ASR decoder attends to the encodings.
However, they have an issue handling a large number of biasing words.
To efficiently handle them, a prefix tree, or a trie, -based deep biasing methods \cite{Le21-CSE, Le21-DSF, Sun21-TCP} have been considered.
A prefix tree of biasing words is built, whose nodes represents a subword symbol.
Given a previously decoded subword sequence, the tree offers a valid subset of subword symbols that follow them.
The decoder attends only the encodings for the valid subset, which is computationally efficient.
We call this {\it subword-level biasing}, as biasing information are represented as subword units at each decoding step.
In contrast, attention-based methods without prefix trees \cite{Pundak18-DC, Jain20-CR, Sathyendra22-CA} perform {\it word-level biasing}.

Tree-constrained Pointer Generator (TCPGen) \cite{Sun21-TCP, Sun23-GNN, Sun23-CB} further extends prefix-tree-based biasing with a pointer-generator network architecture \cite{See17-GTP}.
The model provides an extra probability distribution that can be directly interpolated with the original ASR output probability distribution, which has an advantage of working with a pre-trained ASR model such as Whisper \cite{Radford22-Whisper} without modifying their architecture \cite{Sun23-CB}.
The probability distribution is derived from attention scores computed with a query based on an ASR's context vector and a key based on node encodings provided by a prefix tree.

As rare biasing words sometimes have pronunciations that are difficult to estimate from text, it is important to provide their pronunciation information as a clue to recognize such words.
This is especially common for ideographic characters such as Japanese kanji.
However, existing prefix-tree-based biasing methods rely solely on their textual representations.
While some existing attention-based deep context methods \cite{Chen19-JGP, Pandey23-PR} utilize pronunciation information, they cannot be directly applied to the prefix-tree-based biasing methods including TCPGen.
As those methods perform {\it subword-level biasing}, phonemes aligned to each subword are required.
This is not trivial because pronunciation is typically defined for the entire word.
Subword pronunciation generally cannot be determined by itself, and its phoneme boundaries are often ambiguous.

In this study, we propose introducing subword-level phoneme-aware encodings for TCPGen.
To obtain alignment from a subword to phonemes, we consider using the attention weights of sequence-to-sequence G2P model \cite{Toshniwal16-G2P, Ploujnikov22-SC} or EM algorithm-based alignment \cite{Jiampojamarn07-AMM, Bisani08-JSM, Novak16-PT}.
We also introduce a phoneme-aware query in the attention computation of TCPGen.
As keys are constructed from the phoneme-aware encodings, it would be preferable for queries to be also explicitly aware of phonemes.
To this end, we train the end-to-end ASR model with auxiliary CTC loss whose target is a phoneme sequence, and the CTC predictions are incorporated into the formulation of query in TCPGen.
Experimental evaluations were conducted on the LibriSpeech corpus and the Corpus of Spontaneous Japanese (CSJ).
We confirmed that our proposed method was effective across both English and Japanese, which are languages with completely different pronunciation systems.

\begin{figure}[t]
\centering
\includegraphics[width=\columnwidth]{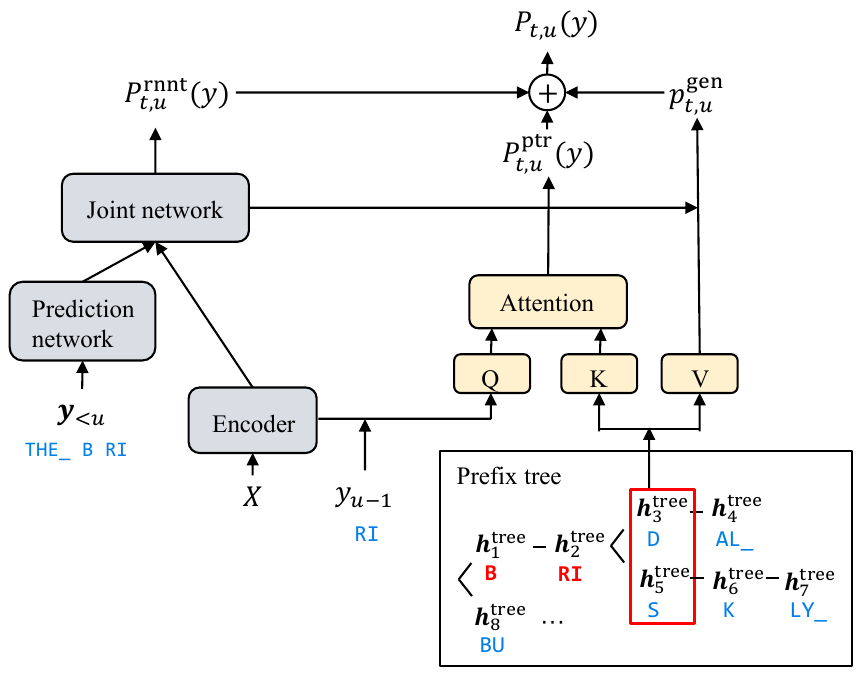}
\caption{Tree-constrained Pointer Generator (TCPGen) for RNN-T.}
\label{fig:tcpgen}
\vspace{-15pt}
\end{figure}

\vspace{-5pt}
\section{Tree-constrained Pointer Generator}
\vspace{-5pt}
\label{sec:tcpgen}
Tree-constrained Pointer Generator (TCPGen) is a prefix-tree-based contextual biasing method for ASR, which has a pointer-generator architecture \cite{Sun21-TCP, Sun23-GNN}.
While the original TCPGen can be applied to both attention-based encoder-decoder \cite{Chorowski15-ABM} and RNN transducer (RNN-T) \cite{Graves12-ST} ASR, we focus on TCPGen for RNN-T in this study, which is illustrated in Fig. \ref{fig:tcpgen}.

A prefix tree of size $N$ is constructed from the biasing word list, where the $n$-th node $(n = 1, ... N)$ corresponds to the $s(n)$-th subword symbol in the vocabulary.
Note that $s(n) \in \{1, ... V\}$ denotes a mapping function from a node index to a vocabulary index, where $V$ is a vocabulary size.
The prefix tree provides a set of active node indices $\mathcal{N}(\bm{y})$ given prefix $\bm{y}$.
TCPGen provides an extra probability distribution to bias the RNN-T output distribution.
The TCPGen probability distribution is noted as $P^{\rm ptr}_{t,u}(y)$ for time step $t$ and RNN-T prediction step $u$, which is calculated via an attention mechanism.
The query for the attention is defined as:
\begin{align}
\label{eq:query}
\bm{q}_{t,u} = W^{\rm q} \bm{h}^{\rm enc}_t + W^{\rm q'} {\rm Emb}(y_{u-1}),
\end{align}
where $\bm{h}^{\rm enc}_t$ and ${\rm Emb}({y}_{u-1})$ denote an encoder output and a previous token embedding, respectively.
Let $\bm{h}^{\rm tree}_n$ denote the encoding of prefix tree node $n$, which is computed with graph neural network (GNN) as described below.
The key and value are generated from $\bm{h}^{\rm tree}_n$ of active nodes $n \in \mathcal{N}(\bm{y}_{<u})$.
For the $s(n)$-th subword symbol, they are defined as:
% The key and value are defined for the $s(n)$-th subword symbol where $n \in \mathcal{T}(\bm{y}_{<u})$ as:
\begin{align}
\label{eq:key-value}
\bm{k}_{s(n)} = W^{\rm k} \bm{h}^{\rm tree}_n, \, \bm{v}_{s(n)} = W^{\rm v} \bm{h}^{\rm tree}_n.
\end{align}
By using Eq. (\ref{eq:query}) and (\ref{eq:key-value}), the TCPGen distribution is denoted as:
\begin{align}
\label{eq:prob-ptr}
P^{\rm ptr}_{t,u}(y) &= {\rm Softmax}(\mask_{s(n), n \in \mathcal{N}(\bm{y}_{<u})}(\bm{q}_{t,u} K^{T} / \sqrt{d})),
\end{align}
where $K$ represents a key matrix, whose $j$-th row corresponds to $\bm{k}_j$ ($1 \leq j \leq V$).
$\mask$ in Eq. (\ref{eq:prob-ptr}) is a function that masks out all but rows for valid vocabulary indices $s(n)$ derived from $\mathcal{N}(\bm{y}_{<u})$.

We used graph convolutional network (GCN) \cite{Kipf17-GCN} of $L$ layers as GNN to compute $\bm{h}^{\rm tree}_n$, following \cite{Sun23-GNN}.
Let $H^{(l)} \in \mathbb{R}^{N \times d}$ denote node encodings at the $l$-th layer.
$\bm{h}^{\rm tree}_n$ corresponds to the $n$-th row of $H^{(L)}$.
$H^{(0)}$ is initialized as:
\begin{align}
\label{eq:encoding}
H^{(0)} = [{\rm Emb}(s(1)), ..., {\rm Emb}(s(n)), ..., {\rm Emb}(s(N))].
\end{align}
$H^{(l)}$ is computed as:
\begin{align}
H^{(l)} = {\rm Relu}(\tilde{D}^{-\frac{1}{2}}\tilde{A}\tilde{D}^{-\frac{1}{2}} H^{(l-1)} W^{(l-1)}),
\end{align}
where $\tilde{A}$ is an adjacency matrix including self-loops, and $\tilde{D}$ is the degree matrix of $\tilde{A}$.

TCPGen also provides a generation probability, or an interpolation weight between RNN-T and the TCPGen probability distribution as:
\begin{align}
\label{eq:prob-gen}
p^{\rm gen}_{t,u} &= \sigma (W^{\rm gen} [\bm{h}^{\rm joint}_{t,u}; \bm{h}^{\rm ptr}_{t,u}]),
\end{align}
where $\bm{h}^{\rm joint}_{t,u}$ denotes the output of RNN-T joint network, and $\bm{h}^{\rm ptr}_{t,u}$ is denoted as:
\begin{align}
\bm{h}^{\rm ptr}_{t,u} = \sum_{j=1}^V P^{\rm ptr}_{t,u}(y = j)\bm{v}_j.
\end{align}
The final output probability distribution is calculated as:
\begin{align}
P_{t,u}(y) = (1 - p^{\rm gen}_{t,u})P^{\rm rnnt}_{t,u}(y) + p^{\rm gen}_{t,u}P^{\rm ptr}_{t,u}(y),
\end{align}
where $P^{\rm rnnt}_{t,u}(y)$ denotes the RNN-T output distribution.

\vspace{-5pt}
\section{Proposed method}
\vspace{-5pt}
\subsection{Phoneme-aware encoding}
\vspace{-5pt}
\label{sec:paware-encoding}
In this study, we investigate a way to leverage pronunciations of biasing words in TCPGen.
Rare words sometimes have pronunciations that are not obvious from their textual form (e.g. choir /\textipa{kwa\textsci \textschwa r}/ and Greenwich /\textipa{gren\textsci t\textesh}/).
This is more likely in Japanese, as it has kanji characters that represent not their pronunciations but their concepts (ideogram).
By feeding pronunciation information as phonemes, we expect the model to align biasing words to speech better and to generalize well to unseen words with unusual pronunciations.
Some contextual biasing studies have shown the effectiveness of using phonemes \cite{Chen19-JGP, Pandey23-PR}, but they are limited to word-level biasing methods, where word-level phoneme representations are employed.

\begin{figure}[t]
\centering
\includegraphics[width=\columnwidth]{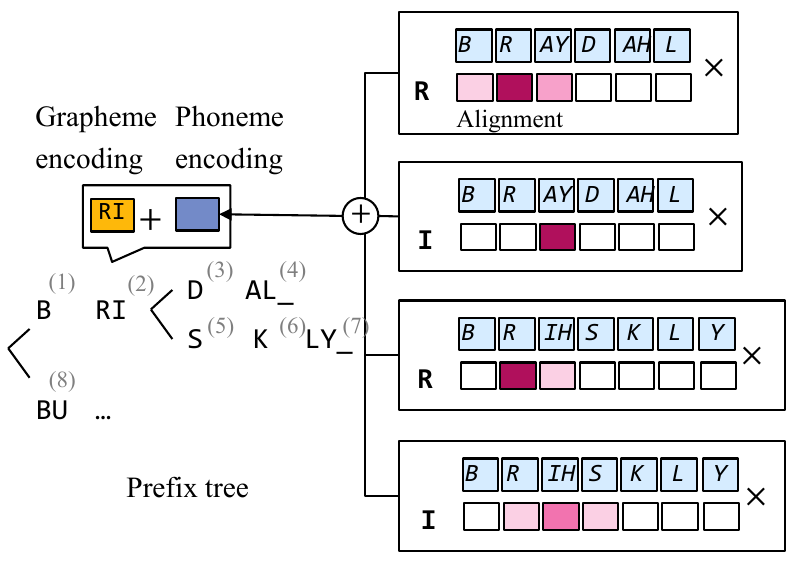}
\caption{The computation of phoneme-aware encoding. For the second node, phoneme embeddings weighted by alignments for (R, BRIDAL), (I, BRIDAL), (R, BRISKLY), and (I, BRISKLY) are summed up.}
\label{fig:encoding}
\vspace{-15pt}
\end{figure}

To leverage phoneme information for a prefix-tree-based method like TCPGen, subword-level phoneme representations that correspond to prefix tree nodes are required.
As the correspondence between subwords and characters is obvious, we here consider how to align characters to phonemes.
We investigate two options for alignment.
One is attention weights of sequence-to-sequence G2P model \cite{Toshniwal16-G2P, Ploujnikov22-SC}, which converts a character sequence into a phoneme sequence.
The other is EM algorithm-based alignment that allows many-to-many character-to-phoneme alignment \cite{Jiampojamarn07-AMM, Bisani08-JSM, Novak16-PT}.
The mapping probabilities are estimated with the forward--backward algorithm, and the Viterbi algorithm is used to obtain the most likely path.
It is worth noting that attention weights provide soft alignment, while EM-based alignment is a hard ($0$ or $1$) and monotonic one.

We here formulate alignments and subword-level phoneme-based encodings for each individual word.
Suppose that word $w$ is represented as a $l_{\rm p}^{(w)}$-length phoneme sequence, a $l_{\rm c}^{(w)}$-length character sequence, and a $l_{\rm s}^{(w)}$-length subword sequence.
We omit superscript $(w)$ from them for simplicity.
Let $P^{(w)} \in \mathbb{R}^{l_{\rm p} \times d_{\rm p}}$ denote an embedding matrix composed of $d_{\rm p}$-dimensional vectors for the phoneme sequence.
Let $A_{\rm c \to p}^{(w)} \in \mathbb{R}^{l_{\rm c} \times l_{\rm p}}$ and $A_{\rm s \to c}^{(w)} \in \mathbb{R}^{{l_{\rm s}} \times l_{\rm c}}$ denote character--phoneme alignments and subword--character alignments, respectively, where $[A_{\rm c \to p}^{(w)}]_{i,j} \in [0,1]$ and $[A_{\rm s \to c}^{(w)}]_{i,j} \in \{0,1\}$.
By using the alignments, the phoneme-based encoding for the $i$-th subword of word $w$ is represented as:
\begin{align}
\label{eq:alignment}
[A_{\rm s \to \rm c}^{(w)}A_{\rm c \to \rm p}^{(w)}P^{(w)}]_i,
\end{align}
where $[\cdot]_i$ denotes the $i$-th row of the matrix.

We then formulate an encoding for each node in a prefix tree.
The tree node corresponds to a subword symbol, which can be shared among two or more words.
For example, in Fig. \ref{fig:tcpgen}, RI is shared between BRIDAL and BRISKLY.
% Let $\mathcal{W}_n$ denotes a word set passing through node $n$.
Let $\mathcal{W}(n)$ denote a word set obtained by traversing from node $n$ to leaf nodes.
The encoding of node $n$ is represented by the summation of ones for all the words $w \in \mathcal{W}(n)$ as:
\begin{align}
\label{eq:pencoding-each}
\bm{e}(n) = W^{\rm p} \sum_{w \in \mathcal{W}(n)} [A_{\rm s \to \rm c}^{(w)}A_{\rm c \to \rm p}^{(w)}P^{(w)}]_{i_n},
\end{align}
where $i_n$ denotes the depth of node $n$.
$W^{\rm p}$ denotes a linear layer that maps $d_{\rm p}$ to $d$, so as to match the size of $\bm{e}(\cdot)$ and that of ${\rm Emb}(\cdot)$.
Finally, TCPGen uses the sum of the phoneme-based encodings and the original grapheme-based encodings in Eq. (\ref{eq:encoding}) as:
\begin{align}
\label{eq:pencoding}
H^{(0)} = [{\rm Emb}(s(1)) + \bm{e}(1), ..., {\rm Emb}(s(N)) + \bm{e}(N)].
\end{align}

Fig. \ref{fig:encoding} shows an example of this encoding computation.
The second ($n=2$) node of the prefix tree corresponds to subword RI, which is shared by the words BRIDAL and BRISKLY.
RI comprises two characters R and I.
For each character and word pair (R, BRIDAL), (I, BRIDAL), (R, BRISKLY), (I, BRISKLY), we obtain the weighted sum of the embeddings of its phoneme sequence, by using character--phoneme alignments as weights.
They are summed up to make the node encoding $\bm{e}(2)$, as in Eq. (\ref{eq:pencoding-each}).

\vspace{-5pt}
\subsection{Phoneme-aware query}
\vspace{-5pt}
\label{sec:paware-query}
The phoneme-aware encodings in Section \ref{sec:paware-encoding} are introduced in the keys of TCPGen in Eq. (\ref{eq:key-value}).
As dot-product attention in Eq. (\ref{eq:prob-ptr}) calculates the relevance between queries and keys, queries should also contains phoneme information to exploit the phoneme-aware encodings.
For this purpose, we add an auxiliary CTC objective on the top of the ASR encoder \cite{Kim17-JCTC}, whose target is a phoneme sequence.
Subsequently, we propose adding the CTC predictions to the query.
Eq. (\ref{eq:query}) is replaced with:
\begin{align}
\bm{q}_{t,u} = W^{\rm q} (\bm{h}^{\rm enc}_t + W^{\rm p}\bm{h}^{\rm ctc}_t) + W^{\rm q'} {\rm Emb}(y_{u-1}),
\end{align}
\vspace{-10pt}
\begin{align}
\bm{h}^{\rm ctc}_t = \argmax_{v \neq \phi} W^{\rm ctc}\bm{h}^{\rm enc}_t,
\end{align}
where $\bm{h}^{\rm ctc}_t$ denotes the embedding of the non-blank phoneme token with the highest probability at time $t$.

\vspace{-5pt}
\section{Experimental evaluations}
\vspace{-5pt}
\subsection{Experimental conditions}
\vspace{-5pt}
We conducted contextual biasing experiments on the LibriSpeech corpus \cite{Panayotov-15LS}.
We followed the author's TCPGen implementation on ESPnet2 \cite{Watanabe18-ESPnet}.
We trained models on the $100$-hour subset and the full set ($960$ hours) of LibriSpeech.
An artificial biasing list was created for each utterance, where rare words were extracted from the reference text, and many distractors were added.
During evaluation, the biasing lists defined in \cite{Le21-CSE} were used.
Rare words are defined as words that fall outside the $5000$ most common words, and $1000$ distractors randomly sampled from rare word set are added.
The text is tokenized into subwords using Byte Pair Encoding (BPE) \cite{Sennrich16-BPE} of vocabulary size $600$.
We treated word boundary as suffix instead of prefix (e.g. "THE\_" instead of "\_THE") as in \cite{Sun21-TCP}.

For ASR, an RNN-T model was used in this study.
The encoder has a Conformer architecture of $12$ layers, $256$ output size, and $4$ attention heads.
The prediction network consists of a single layer LSTM of size $256$ or $512$, and the joint network consists of a single linear layer of size $320$.
We trained a baseline ASR model for $70$ epochs for the $100$h subset and $25$ epochs for the full set, with Adam optimizer of learning rate $0.002$ and $15000$ warmup steps.
We applied speed perturbation and SpecAugment with adaptive masking \cite{Park19-SA}.
The decoding is done with beam size $20$.
For contextual biasing, we added a TCPGen architecture upon the baseline ASR and jointly trained them.
During its training on the $100$h data, biasing lists are created on the fly by adding $500$ distractors, where rare words are defined as words that appears less than $15$ times in the training data, following \cite{Sun21-TCP}.
On the $960$h data, the threshold was set to $120$ (times).
We used $6$-layer GCN encodings as described in Section \ref{sec:tcpgen}, whose size $d$ was the same as that of the prediction network ($256$ or $512$).
We trained the model for $30$ epochs without TCPGen for the $100$h subset and $10$ epochs for the full set.

For proposed phoneme-aware encoding, we used SoundChoice \cite{Ploujnikov22-SC} or Phonetisaurus \cite{Novak16-PT} for G2P.
For SoundChoice G2P, we used the SpeechBrain model \footnote{https://huggingface.co/speechbrain/soundchoice-g2p} trained on the LibriG2P training data \footnote{https://huggingface.co/datasets/flexthink/librig2p-nostress-space}.
We obtained the character--phoneme alignment from its attention weight.
For Phonetisaurus G2P, we trained it on the LibriSpeech data and calculated EM-based alignment.
The phoneme vocabulary consists of $39$ unique symbols.

We also conducted experiments on the Corpus of Spontaneous Japanese (CSJ) \cite{maekawa03-CSJ}.
We used the CSJ-SPS subset ($280$ hours) for training, which is the recordings of simulated public speaking on everyday topics.
We used the eval1 set for evaluation, which is the recordings of academic lectures that includes many uncommon terms.
We created the biasing list in the same way as LibriSpeech, where rare words are defined as words that are not in the $2000$ most common words (which accounts for $90\%$ of all the word occurrences) in the CSJ-SPS data.
We followed the LibriSpeech configuration for ASR and TCPGen.
For G2P, we regarded the G2P conversion results from pyopenjtalk \footnote{https://github.com/r9y9/pyopenjtalk} as the reference.
We trained SoundChoice and Phonetisaurus G2P from scratch for them.
Pyopenjtalk cannot be directly used for our purpose, as it does G2P conversion based on dictionary, where the alignment cannot be obtained.
The BPE and phoneme vocabulary sizes were 5000 and 39.

\vspace{-5pt}
\subsection{Experimental results}
\vspace{-5pt}

\begingroup
\renewcommand{\arraystretch}{1.1}
\begin{table}[t]
\caption{The comparison of encoding for TCPGen. The models are trained on LibriSpeech $100$h. ``TCPGen+p'' means TCPGen with phoneme-aware encoding, which is our proposal. ``Align'' denotes a way to get alignment (Att: attention, EM: EM algorithm). ``Pemb'' denotes a way to get phoneme embeddings (G2P: G2P, OH: one-hot, OH+: one-hot + linear layer). $d$ was set to $512$ to match with G2P's.}
\vspace{3pt}
\label{tab:encoding}
\centering
\begin{tabular}{lccc} \hline
 & Grapheme & Phoneme & WER/RWER \\
 &  & (Align/Pemb) & clean \\ \hline
(\url{a1}) No bias & - & - & $6.8/27.3$ \\
(\url{a2}) TCPGen & \checkmark & - & $5.4/15.0$ \\
(\url{a3}) TCPGen+p & \checkmark & \checkmark(Att/G2P) & $5.4/14.2$ \\
(\url{a4}) TCPGen+p & \checkmark & \checkmark(Att/OH) & $5.4/14.9$ \\
(\url{a5}) TCPGen+p & \checkmark & \checkmark(Att/OH+) & $\bm{5.3}/\bm{14.0}$ \\
(\url{a6}) TCPGen+p & - & \checkmark(Att/OH+) & $5.8/18.0$ \\
(\url{a7}) TCPGen+p & \checkmark & \checkmark(EM/OH+) & $5.4/14.3$ \\
\hline
\end{tabular}
\vspace{-10pt}
\end{table}
\endgroup

\begingroup
\renewcommand{\arraystretch}{1.1}
\begin{table}[t]
\caption{The effect of phoneme-aware queries. The models are trained on LibriSpeech $100$h. The ``CTC MTL'' column denotes the target unit of CTC mulitask learning (sw: subword, phn: phoneme). ``phn-aware Q'' denotes phone-aware querying. $d$ was set to $256$.}
\vspace{3pt}
\label{tab:pquery}
\centering
\begin{tabular}{lcc} \hline
 & \multicolumn{2}{c}{WER/R-WER} \\
 & clean & other \\\hline
(\url{b1}) No bias & $6.5/26.1$ & $17.9/52.5$ \\ 
(\url{b2}) +CTC (sw) & $6.5/26.7$ & $18.0/53.3$ \\
(\url{b3}) +CTC (phn) & $6.5/26.1$ & $17.9/52.7$ \\ \hline
(\url{b4}) TCPGen & $5.3/14.1$ & $15.9/33.8$ \\
(\url{b5}) +CTC (sw) & $5.1/14.9$ & $15.7/35.0$ \\
(\url{b6}) +CTC (phn) & $5.1/13.6$ & $15.9/34.3$ \\ \hline
(\url{b7}) TCPGen+p & $5.1/13.4$ & $15.9/34.9$ \\ 
(\url{b8}) +CTC (phn) & $5.0/13.1$ & $\bm{15.6}/32.7$ \\
(\url{b9}) ++phn-aware Q & $\bm{4.9}/\bm{12.1}$ & $\bm{15.6}/\bm{31.5}$ \\ \hline
\end{tabular}
\vspace{-10pt}
\end{table}
\endgroup

\begingroup
\renewcommand{\arraystretch}{1.1}
\begin{table}[t]
\caption{The results on LibriSpeech $960$h.}
\vspace{3pt}
\label{tab:libri960}
\centering
\begin{tabular}{lcc} \hline
 & \multicolumn{2}{c}{WER/R-WER} \\
 & clean & other \\ \hline
(\url{c1}) No bias & $3.0/11.8$ & $7.5/26.6$ \\
(\url{c2}) TCPGen & $2.3/4.9$ & $\bm{5.9}/12.9$ \\
(\url{c3}) TCPGen+p +phn-aware Q & $\bm{2.2}/\bm{4.6}$ & $6.0/\bm{12.3}$ \\ \hline
\end{tabular}
\vspace{-5pt}
\end{table}
\endgroup

\begingroup
\renewcommand{\arraystretch}{1.1}
\begin{table}[t]
\vspace{-10pt}
\caption{The result on CSJ (Japanese). The models are trained on the SPS subset of CSJ and evaluated on eval1 set.}
\vspace{3pt}
\label{tab:csj}
\centering
\begin{tabular}{lc} \hline
 & WER/R-WER \\ \hline
(\url{d1}) No bias & $14.3/44.9$ \\ 
(\url{d2}) TCPGen & $13.5/38.8$ \\
(\url{d3}) TCPGen +CTC (phn) & $13.2/37.8$ \\
(\url{d4}) TCPGen+p (Align:Att) & $13.3/37.8$  \\
(\url{d5}) TCPGen+p (Align:Att) +phn-aware Q & $12.6/33.4$  \\
(\url{d6}) TCPGen+p (Align:EM) & $13.1/37.3$ \\ 
(\url{d7}) TCPGen+p (Align:EM) +phn-aware Q & $\bm{12.2}/\bm{30.6}$ \\ \hline
\end{tabular}
\vspace{-18pt}
\end{table}
\endgroup

First, we compared different ways to compute encodings for TCPGen trained on the LibriSpeech $100$-hour subset in Table \ref{tab:encoding}.
As evaluation metrics, we used a rare word error rate (R-WER), along with a word error rate (WER).
The R-WER was calculated only on rare words included in biasing lists, as defined in \cite{Le21-CSE, Sun21-TCP}.
As contextual biasing aims to improve the recognition of biasing words in the list without degrading that of words not in the list, our target is to improve the R-WER without degrading the overall WER that includes common words.
In the table, the ``Grapheme'' and ``Phoneme'' column denotes whether to use grapheme-based and phoneme-based node encoding as in Eq. (\ref{eq:encoding}) and (\ref{eq:pencoding}), respectively.
For phoneme-based encoding, we compared different methods to obtain alignments (Align) noted as $A_{\rm c \to \rm p}^{(w)}$ in Eq. (\ref{eq:alignment}) and phoneme embeddings (Pemb) noted as $P^{(w)}$.
The row (\url{a1}) shows the baseline ASR without any contextual biasing.
The row (\url{a2}) shows the existing TCPGen that relies solely on grapheme-based encoding.
In row (\url{a3}) to (\url{a7}), proposed phoneme-aware encoding was introduced to TCPGen, which is denoted as ``TCPGen+p''.
We first used attention weights (denoted as ``Att'') from SoundChoice and compared phoneme embeddings in (\url{a3}) to (\url{a5}).
The row (\url{a3}) uses the G2P model's input embedding of size $512$.
To be comparable to that, the size of encoding $d$ was set to $512$ in Table \ref{tab:encoding}.
We used one-hot representations in (\url{a4}) and (\url{a5}), where the row (\url{a4}) used them as they are (``OH'') while the row (\url{a5}) added a linear layer (``OH+''), as noted $W^{\rm p}$ in Eq. (\ref{eq:pencoding-each}).
Note that the linear layer in (\url{a5}) increases the number of parameters by $0.03$M, which is relatively much smaller to the entire model ($0.1\%$).
Among them, one-hot representations with a linear layer (\url{a5}) performed best.
This TCPGen with phoneme-aware encoding achieved better R-WER and WER than existing TCPGen (\url{a2}).
We explored to use phoneme-based encoding only, without grapheme-based encoding (\url{a6}), but this degraded the performance.
We also explored to use EM-based alignment (\url{a7}), but it was not better than attention-based alignment.

Secondly, we saw the effect of phoneme-aware queries on the $100$h subset.
The results are listed in Table \ref{tab:pquery}.
We used the best performing encoding with attention-based alignment and one-hot representations with a linear mapping, found in Table \ref{tab:encoding}.
Note that $d$ was set to $256$ in Table \ref{tab:pquery}.
As described in Section \ref{sec:paware-query}, we added CTC loss in addition to RNN-T loss, whose weight was set to $0.3$.
The target of CTC loss was subword (sw) or phoneme (phn) unit.
We found using phoneme-aware queries (denoted as ``phn-aware Q'') (\url{b9}) further improved the performance of TCPGen with phoneme-aware encoding (``TCPGen+p'') (\url{b7}).
The baselines with multi-task learning (\url{b2}), (\url{b3}), (\url{b5}), and (\url{b6}) did not show such a significant improvement, indicating that the improvement was owing not to multi-task learning itself but to our proposed method.
Table \ref{tab:libri960} shows the results on the LibriSpeech full ($960$ hours) set.
We found TCPGen with our proposed method (phoneme-aware encoding and phone-aware querying) outperformed the existing TCPGen in terms of the R-WER, with a minimum degradation in the WER.

Table \ref{tab:csj} shows the results on the Japenese corpus CSJ.
On the Japanese corpus, EM-based alignment (``Align:EM'') (\url{d6}) and (\url{d7}) was better than attention-based one (``Align:Att'') (\url{d4}) and (\url{d5}).
We assume that, as Japanese syllables are pronounced distinctly, hard alignment with EM algorithm works better.
On the other hand, in English, blending of sounds between syllables occurs, and therefore soft alignment would be better, as shown in Table \ref{tab:encoding}.
We found, with phoneme-aware queries (\url{d7}), the WER/R-WER was significantly improved to $12.2/30.6$, which was significantly better than the existing TCPGen (\url{d2}) and its CTC multi-task learning version (\url{d3}).

\vspace{-12pt}
\section{Conclusions}
\vspace{-8pt}
In this study, we propose introducing phoneme-aware encoding to a prefix-tree-based contextual biasing method TCPGen.
We align phoneme representations to subword-level tree nodes using alignment from G2P attention weights or the EM algorithm.
We further propose incorporating CTC predictions of phonemes into the queries of TCPGen, so as to better understand phoneme-aware encodings in its keys.
We have experimentally shown that TCPGen with our proposed phoneme-aware encoding consistently improved the contextual biasing performance on both LibriSpeech and CSJ.

\bibliographystyle{IEEEbib}
\bibliography{strings,refs}

\begin{thebibliography}{10}

\bibitem{Prabhavalkar23-E2E}
Rohit Prabhavalkar, Takaaki Hori, Tara~N. Sainath, Ralf Schluter, and Shinji Watanabe,
\newblock ``End-to-end speech recognition: A survey,''
\newblock {\em ArXiv}, 2023.

\bibitem{Zhao19-SF}
Ding Zhao, Tara~N. Sainath, David Rybach, Pat Rondon, Deepti Bhatia, Bo~Li, and Ruoming Pang,
\newblock ``Shallow-fusion end-to-end contextual biasing,''
\newblock in {\em Interspeech}, 2019.

\bibitem{Gourav21-PS}
Aditya Gourav, Linda Liu, Ankur Gandhe, Yile Gu, Guitang Lan, Xiangyang Huang, Shashank Kalmane, Gautam Tiwari, Denis Filimonov, Ariya Rastrow, Andreas Stolcke, and Ivan Bulyko,
\newblock ``Personalization strategies for end-to-end speech recognition systems,''
\newblock in {\em ICASSP}, 2021.

\bibitem{Pundak18-DC}
Golan Pundak, Tara~N. Sainath, Rohit Prabhavalkar, Anjuli Kannan, and Ding Zhao,
\newblock ``Deep context: End-to-end contextual speech recognition,''
\newblock {\em SLT}, 2018.

\bibitem{Jain20-CR}
Mahaveer Jain, Gil Keren, Jay Mahadeokar, and Yatharth Saraf,
\newblock ``Contextual {RNN-T} for open domain {ASR},''
\newblock in {\em Interspeech}, 2020.

\bibitem{Sathyendra22-CA}
Kanthashree~Mysore Sathyendra, Thejaswi Muniyappa, Feng-Ju Chang, Jing Liu, Jinru Su, Grant~P. Strimel, Athanasios Mouchtaris, and Siegfried Kunzmann,
\newblock ``Contextual adapters for personalized speech recognition in neural transducers,''
\newblock {\em ICASSP}, 2022.

\bibitem{Le21-CSE}
Duc Le, Mahaveer Jain, Gil Keren, Suyoun Kim, Yangyang Shi, Jay Mahadeokar, Julian Chan, Yuan Shangguan, Christian Fuegen, Ozlem Kalinli, Yatharth Saraf, and Michael~L. Seltzer,
\newblock ``Contextualized streaming end-to-end speech recognition with trie-based deep biasing and shallow fusion,''
\newblock in {\em Interspeech}, 2021.

\bibitem{Le21-DSF}
Duc Le, Gil Keren, Julian Chan, Jay Mahadeokar, Christian Fuegen, and Michael~L. Seltzer,
\newblock ``Deep shallow fusion for {RNN-T} personalization,''
\newblock in {\em SLT}, 2021.

\bibitem{Sun21-TCP}
Guangzhi Sun, Chao Zhang, and Philip~C. Woodland,
\newblock ``Tree-constrained pointer generator for end-to-end contextual speech recognition,''
\newblock {\em ASRU}, 2021.

\bibitem{Sun23-GNN}
Guangzhi Sun, C.~Zhang, and Phil Woodland,
\newblock ``Graph neural networks for contextual {ASR} with the tree-constrained pointer generator,''
\newblock {\em CoRR}, 2023.

\bibitem{Sun23-CB}
Guangzhi Sun, Xianrui Zheng, Chao Zhang, and Philip~C. Woodland,
\newblock ``Can contextual biasing remain effective with {Whisper} and {GPT-2}?,''
\newblock in {\em Interspeech}, 2023.

\bibitem{See17-GTP}
Abigail See, Peter~J. Liu, and Christopher~D. Manning,
\newblock ``Get to the point: Summarization with pointer-generator networks,''
\newblock in {\em ACL}, 2017.

\bibitem{Radford22-Whisper}
Alec Radford, Jong~Wook Kim, Tao Xu, Greg Brockman, Christine McLeavey, and Ilya Sutskever,
\newblock ``Robust speech recognition via large-scale weak supervision,''
\newblock {\em ArXiv}, 2022.

\bibitem{Chen19-JGP}
Zhehuai Chen, Mahaveer Jain, Yongqiang Wang, Michael~L. Seltzer, and Christian Fuegen,
\newblock ``Joint grapheme and phoneme embeddings for contextual end-to-end {ASR},''
\newblock in {\em Interspeech}, 2019.

\bibitem{Pandey23-PR}
Rahul Pandey, Roger Ren, Qi~Luo, Jing Liu, Ariya Rastrow, Ankur Gandhe, Denis Filimonov, Grant Strimel, Andreas Stolcke, and Ivan Bulyko,
\newblock ``Procter: Pronunciation-aware contextual adapter for personalized speech recognition in neural transducers,''
\newblock in {\em ICASSP}, 2023.

\bibitem{Toshniwal16-G2P}
Shubham Toshniwal and Karen Livescu,
\newblock ``Jointly learning to align and convert graphemes to phonemes with neural attention models,''
\newblock in {\em SLT}, 2016.

\bibitem{Ploujnikov22-SC}
Artem Ploujnikov and Mirco Ravanelli,
\newblock ``{SoundChoice}: Grapheme-to-phoneme models with semantic disambiguation,''
\newblock in {\em Interspeech}, 2022.

\bibitem{Jiampojamarn07-AMM}
Sittichai Jiampojamarn, Grzegorz Kondrak, and Tarek Sherif,
\newblock ``Applying many-to-many alignments and hidden markov models to letter-to-phoneme conversion,''
\newblock in {\em NAACL}, 2007.

\bibitem{Bisani08-JSM}
Maximilian Bisani and Hermann Ney,
\newblock ``Joint-sequence models for grapheme-to-phoneme conversion,''
\newblock {\em Speech Communication}, 2008.

\bibitem{Novak16-PT}
Josef~Robert Novak, Nobuaki Minematsu, and Keikichi Hirose,
\newblock ``Phonetisaurus: Exploring grapheme-to-phoneme conversion with joint n-gram models in the {WFST} framework,''
\newblock {\em Natural Language Engineering}, 2016.

\bibitem{Chorowski15-ABM}
Jan~K Chorowski, Dzmitry Bahdanau, Dmitriy Serdyuk, Kyunghyun Cho, and Yoshua Bengio,
\newblock ``Attention-based models for speech recognition,''
\newblock in {\em NIPS}, 2015.

\bibitem{Graves12-ST}
Alex Graves,
\newblock ``Sequence transduction with recurrent neural networks,''
\newblock {\em ArXiv}, 2012.

\bibitem{Kipf17-GCN}
Thomas~N. Kipf and Max Welling,
\newblock ``Semi-supervised classification with graph convolutional networks,''
\newblock in {\em ICLR}, 2017.

\bibitem{Kim17-JCTC}
Suyoun Kim, Takaaki Hori, and Shinji Watanabe,
\newblock ``Joint {CTC}-attention based end-to-end speech recognition using multi-task learning,''
\newblock {\em ICASSP}, 2017.

\bibitem{Panayotov-15LS}
Vassil Panayotov, Guoguo Chen, Daniel Povey, and Sanjeev Khudanpur,
\newblock ``Librispeech: An {ASR} corpus based on public domain audio books,''
\newblock in {\em ICASSP}, 2015.

\bibitem{Watanabe18-ESPnet}
Shinji Watanabe, Takaaki Hori, Shigeki Karita, Tomoki Hayashi, Jiro Nishitoba, Yuya Unno, Nelson {Enrique Yalta Soplin}, Jahn Heymann, Matthew Wiesner, Nanxin Chen, Adithya Renduchintala, and Tsubasa Ochiai,
\newblock ``{ESPnet}: End-to-end speech processing toolkit,''
\newblock in {\em Interspeech}, 2018.

\bibitem{Sennrich16-BPE}
Rico Sennrich, Barry Haddow, and Alexandra Birch,
\newblock ``Neural machine translation of rare words with subword units,''
\newblock in {\em ACL}, 2016.

\bibitem{Park19-SA}
Daniel~S. Park, Yu~Zhang, Chung-Cheng Chiu, Youzheng Chen, Bo~Li, William Chan, Quoc~V. Le, and Yonghui Wu,
\newblock ``Specaugment on large scale datasets,''
\newblock {\em ICASSP}, 2019.

\bibitem{maekawa03-CSJ}
K.~Maekawa,
\newblock ``{Corpus of Spontaneous Japanese} : its design and evaluation,''
\newblock {\em SSPR}, 2003.

\end{thebibliography}

\end{document}